\documentclass[sn-vancouver,iicol]{sn-jnl}

\usepackage[protrusion=true,expansion=true]{microtype}
\usepackage{array}
\usepackage{xfrac}
\usepackage[caption=false, captionskip=0pt, farskip=0pt]{subfig}
\usepackage{pifont}

\usepackage{rotating}
\usepackage{makecell}
\usepackage{multirow}
\usepackage{verbatim}
\usepackage{hyperref}
\hypersetup{
  pdftitle={Quantifying the Effect of Feedback Frequency in Interactive Reinforcement Learning for Robotic Tasks},
  pdfauthor={Daniel Harnack, Julie Pivin-Bachler, and Nicolás Navarro-Guerrero},
  pdfsubject={Robotics (cs.RO); Artificial Intelligence (cs.AI), Human-Computer Interaction (cs.HC)},
  pdfkeywords={Interactive Reinforcement Learning, Human-aligned Reinforcement Learning, Guided Exploration, Intrinsic Feedback Homology},%
}

\jyear{2022}%

\begin{document}

\title[Effect of Feedback Frequency in IRL Performance]{Quantifying the Effect of Feedback Frequency in Interactive Reinforcement Learning for Robotic Tasks}

\author[1]{\fnm{Daniel} \sur{Harnack}}\email{daniel.harnack@dfki.de}

\author[2]{\fnm{Julie} \sur{Pivin-Bachler}}\email{julie.pivin-bachler@univ-tlse3.fr}

\author*[1]{\fnm{Nicol\'as} \sur{Navarro-Guerrero}}\email{nicolas.navarro@dfki.de}

\affil*[1]{\orgdiv{Robotics Innovation Center}, \orgname{Deutsches Forschungszentrum für Künstliche Intelligenz (DFKI) GmbH}, \orgaddress{\street{Robert-Hooke-Stra{\ss}e 1}, \city{Bremen}, \postcode{28359}, \state{Bremen}, \country{Germany}}}

\affil[2]{\orgdiv{Robotics and Interactive Systems -- UPSSITECH}, \orgname{University Paul Sabatier}, \orgaddress{\street{118 Route de Narbonne}, \city{Toulouse}, \postcode{31062}, \state{Occitanie}, \country{France}}}

\abstract{
	Reinforcement learning (RL) has become widely adopted in robot control. Despite many successes, one major persisting problem can be very low data efficiency. One solution is interactive feedback, which has been shown to speed up RL considerably. As a result, there is an abundance of different strategies, which are, however, primarily tested on discrete grid-world and small scale optimal control scenarios. In the literature, there is no consensus about which feedback frequency is optimal or at which time the feedback is most beneficial. To resolve these discrepancies we isolate and quantify the effect of feedback frequency in robotic tasks with continuous state and action spaces. The experiments encompass inverse kinematics learning for robotic manipulator arms of different complexity. We show that seemingly contradictory reported phenomena occur at different complexity levels. Furthermore, our results suggest that no single ideal feedback frequency exists. Rather that feedback frequency should be changed as the agent's proficiency in the task increases.
	}


\maketitle

\section{Introduction}
\label{sec:introduction}
Reinforcement Learning (RL) has become widely used in modern robotic technologies. 
One reason is the compelling simplicity and generality of the framework. 
In short, the behaviour an agent is expected to learn is encoded by a reward function. 
Through interaction with the environment, the agent will learn to maximize the reward by performing actions that have proven to be beneficial. However, this seeming simplicity has many pitfalls and subtleties. 
One common shortcoming of most algorithms is the very low data efficiency: complex problems might require millions of agent-environment interactions to be solved \cite[e.g.,][]{Silver2017Mastering}.

\begin{figure}[ht]
    \centering
    \includegraphics[width=\columnwidth]{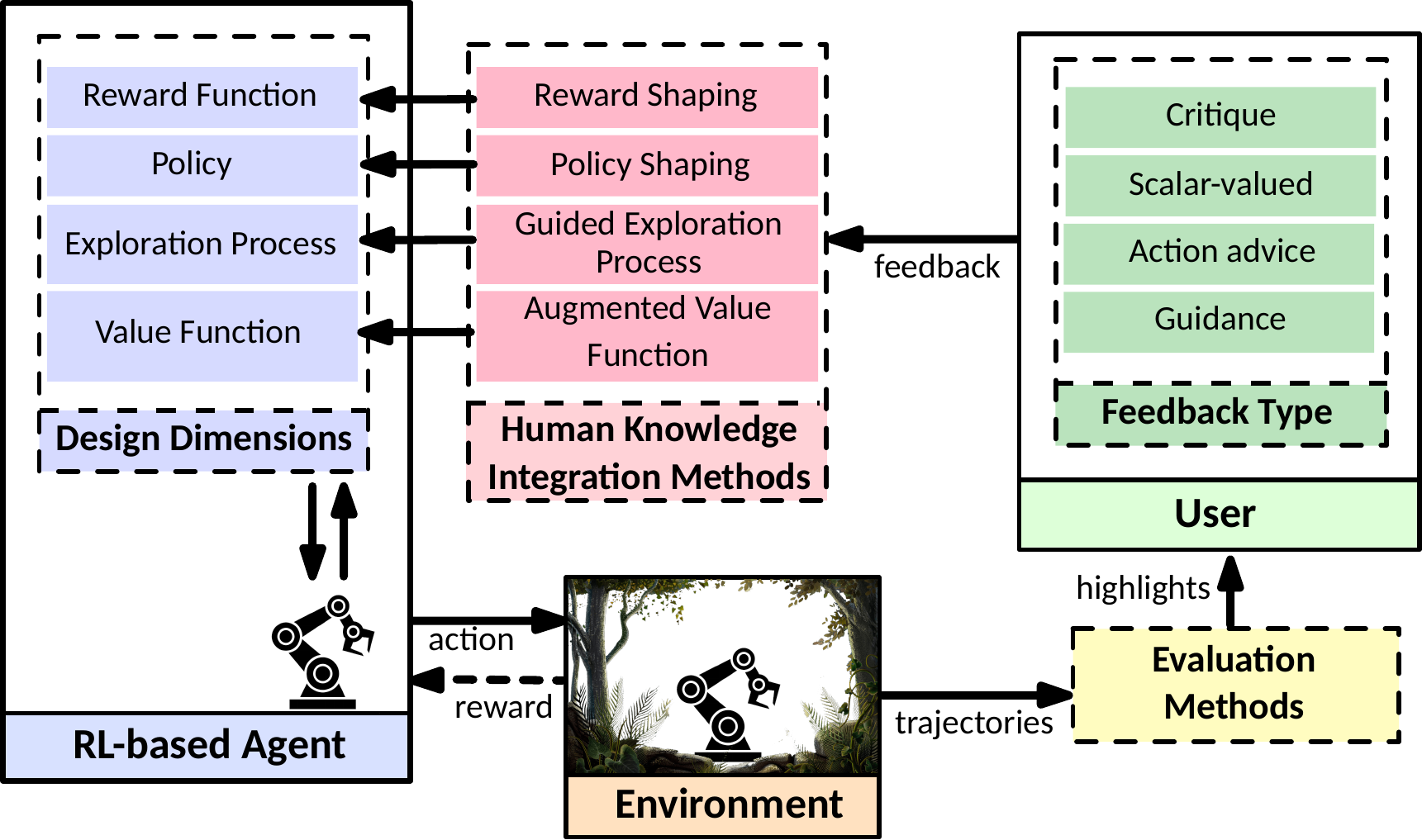}
    \caption{Summary of alternative implementations of interactive reinforcement learning. Adapted from \cite{ArzateCruz2020Survey}.}
    \label{fig:irl}
\end{figure}

One strategy to accelerate this procedure is interactive reinforcement learning (IRL) \cite{ArzateCruz2020Survey}. Interaction augments the sources of information provided to the learning agents by teacher feedback, which can be a human or another type of agent \cite{Tan1997MultiAgent}. 
In the latter case, it is also known as the \textit{agents teaching agents} subfield of transfer learning \cite{DaSilva2019Agents}.
There are numerous alternatives to implement IRL as described by Arzate Cruz and Igarashi~\cite{ArzateCruz2020Survey}. A graphical overview is provided in Figure~\ref{fig:irl}. Firstly, the teacher feedback can be classified into critique (binary), scalar values, action advice and guidance. Further, this feedback can be used to modify different aspects of the learning model, i.e., the reward function (reward shaping \cite[eg.,][]{Ng1999Policy}), the policy (policy-shaping \cite[eg.,][]{Griffith2013Policy}), the exploration process (guided exploration process \cite[eg.,][]{Stahlhut2015Interactiona}), and the value function (augmented value function).

The literature on interactive reinforcement learning is extensive, such that many combinations shown in Figure~\ref{fig:irl} have been explored already. 
One consistent result is that interaction in any form can perform better than vanilla RL agents. 
However, it is still unclear how much the different aspects contribute to overall performance gains or how different feedback strategies can be combined, and in which proportions, to optimize users' experience, agents' task performance, or both.

IRL algorithms are mainly developed with human teachers in mind. 
Empirical evidence indicates that people are strongly biased to use evaluative feedback communicatively rather than as reinforcement \cite{Ho2019People, Thomaz2008Teachable}. In other words, humans use \textit{evaluative feedback as communication} as a policy-shaping strategy rather than reward shaping.
Arguably, this type of feedback favours IRL strategies based on \textit{policy shaping} and \textit{guided exploration}, because it would make the interaction both more engaging for the teacher and more effective for the learning agent \cite[eg.,][]{Thomaz2008Teachable, Loftin2014StrategyAware, Griffith2013Policy}. 
Results from Ho et al. \cite{Ho2019People}, showing that people consistently use feedback communicatively even when interacting with reward-maximising agents, further support this claim. 
Moreover, using feedback signals as rewards and punishments when interacting with reward-maximising RL agents can lead to \textit{reward hacking} \cite{Knox2012Reinforcement}. Reward hacking is a consequence of misspecified reward functions, which lead to undesired behaviours, such as when action sequences leading to the reward from the human are repeated at the expense of learning to complete the task more generally. 

Despite efforts to improve the study of interaction on RL agents with human teachers, we believe there is still much to be learned using simulated teachers and oracles as suggested by Bignold et al.\ \cite{Bignold2021Evaluation}. 
Moreover, human feedback varies in accuracy, availability, concept drift, reward bias, cognitive bias, knowledge level, latency, etc.\ \cite{Bignold2021Evaluation}, which makes it very challenging to isolate the effects of different interaction strategies in reinforcement learning agents. 
Fortunately, pre-trained agents or hard-coded heuristics can be used as feedback sources without requiring modifications to the learning algorithms. These types of `teachers' are primarily used in theoretical research since it allows for better controllable and more easily implementable experiments.

Different strategies have been compared regarding policy-shaping, such as early advising, alternating or stochastic advising, importance advising, and mistake correction. 
Mistake correction consistently outperformed the other methods, both in discrete \cite{Taylor2014Reinforcement, Cruz2017AgentAdvising} and continuous state and action spaces \cite{Taylor2014Reinforcement}. 
Taylor et al.\ \cite{Taylor2014Reinforcement} also noted that mistake correction is more robust to changes in feedback quality than alternate advising. In addition, mistake correction and predictive advising are most robust to changes in the state representation between teacher and agent.
However, as noted by Cruz et al.\ \cite{Cruz2017AgentAdvising}, mistake correction in policy shaping would be the most difficult strategy to implement in real-world scenarios with human teachers since it requires the teacher to detect the mistake, revert it, and suggest a better alternative action. 

A more straightforward way is using mistake correction for guided exploration. Here, the teacher must detect and revert the error but not necessarily suggest a better action. In addition, despite its popularity, we believe that policy-shaping strategies might hinder the learning of robust policies by reducing exploration, which leads to good performance primarily in the neighbourhood of the behaviours demonstrated by the teacher \cite{Suay2011Effect}. Limiting the exploration in this manner can result in poor performance in other areas of the state space not or rarely encountered during training \cite{Suay2011Effect}.

In the literature on feedback-guided exploration, Stahlhut et al.\ \cite{Stahlhut2015Interaction, Stahlhut2015Interactiona} found that mistake correction does not help to increase the learning speed in simple tasks but only starts to have a measurable effect as the task complexity increases. 
It was also observed that higher feedback frequencies lead to more robust agents, i.e., that the average agents' performance for the same hyperparameters has a smaller standard deviation across different random seeds. 
Moreover, feedback can offset the detrimental effect of poorly tuned hyperparameters as a byproduct of this increased robustness.
This effect becomes stronger as feedback frequency increases, regardless of the complexity of the problem. 
Stahlhut et al.\ also speculate that interaction has a more significant impact during early learning. 
In addition, they hypothesise that feedback may be indispensable to achieve sufficient performance in very complex tasks, in agreement with Suay and Chernova's hypothesis \cite{Suay2011Effect}.

Millán-Arias et al.\ \cite{Millan-Arias2020Robust} further investigated the effect of feedback frequency used to guide the exploration process. 
They observed that too much feedback might lead to delayed onset of learning. 
Despite that, the performance of highly interactive agents converges at the same time as the performance of less interactive ones.
In addition, they speculate that too much binary advice, even if 100\% correct, can be counterproductive and slow down learning, particularly in noisy environments. They conclude that intermediate interaction frequencies are optimal.

Summarizing previous findings, there are strikingly contradictory accounts of the optimal choice of feedback frequencies. 
At the same time, some authors suggest that more feedback is better \cite{Stahlhut2015Interaction, Stahlhut2015Interactiona}, while others indicate that the cost of high feedback frequencies does not justify the gains \cite{Bignold2021Evaluation}. 
In contrast, others advocate for intermediate feedback frequencies and report even detrimental effects of frequent feedback \cite{Millan-Arias2020Robust}. 
It was also suggested that the feedback frequency should not be stationary but adjusted as the agents' proficiency increases during training \cite{Cruz2016Training, Stahlhut2015Interaction, Stahlhut2015Interactiona}. 

We assume that the reason for the disagreement is a lack of knowledge about the differential effects of varying feedback frequencies at different levels of agent proficiency and task complexity.
Consolidating these previous findings without further experimentation is complicated since most results only show cumulated reward or sequence length. Moreover, effect sizes, average performance, and statistical analyses are not reported in most cases. Also, the complexity of the setup might make it impossible to isolate the effects of the different parameters used \cite{Cruz2016Training}. In addition, the most common testbeds in IRL research are grid worlds and other low-dimensional discrete state and action spaces. Whereas the small size of these testbeds allows for short experiments, we believe results obtained in those testbeds might not generalize well to more complex problems with real-world implications \cite{ArzateCruz2020Survey}.

Thus, in this paper, we aim to isolate and quantify the effect of feedback frequency on learning performance for different task complexities and agent proficiencies, to shed some light on seemingly contradictory results. As testbeds, we use robotic tasks of varying complexity and continuous action and state spaces. 
We focus on feedback as mistake correction to guide the exploration process since it does not demand expert knowledge of the task. 
In our experiments, we reproduce various seemingly contradictory findings about the optimal choice of interaction frequencies and relate them to a differential effect of the teacher interaction on task complexity. We also show that optimal feedback frequencies typically exhibit temporal drifts, making it difficult to recommend a single range of feedback frequencies for any task. 
We instead conclude, in line with previous suggestions \cite{Cruz2016Training, Stahlhut2015Interaction, Stahlhut2015Interactiona}, that an adaptive interaction regime, which changes with the agents' proficiency, is likely optimal. Finally, we discuss a simple heuristic for choosing a close-to-optimal temporal trajectory for the interaction rate.

\section{Methods}
\label{sec:methods}
This section details all experimental and analysis methods used in the paper.

\subsection{Environment}
\label{sec:world}
Inspired by Stahlhut et al.\ \cite{Stahlhut2015Interaction, Stahlhut2015Interactiona}, we study the effect of feedback frequency as exploration guidance in an inverse kinematic learning task. 
The environment dynamics were implemented by the forward kinematics models of the NAO and KUKA (LBR iiwa 14 R820) robots. 

For the NAO robot, the forward kinematics model described by Kofinas et al.\ \cite{Kofinas2015Complete} was used. Two conditions for the NAO robot were defined: a 2 degrees of freedom (DoFs), and a 4~DoFs condition. 
For the 2~DoFs configuration, the elbow and shoulder roll were actuated.
In the 4~DoF condition, all four joints are used, i.e., shoulder pitch, shoulder roll, elbow yaw and elbow roll.

The KUKA LBR iiwa 14 R820 kinematics were simulated with the model described by Busson et al.\ \cite{Busson2017TaskOriented}. 
For the KUKA arm, three conditions were studied, i.e., 2~DoFs, 4~DoFs, and 7~DoFs conditions. 
For the 2~DoFs configuration, joints 2 and 4 were actuated while keeping the other joints in their respective zero-position. 
For the 4~DoFs configuration, the first four joints were actuated while keeping the other joints in their respective zero-position, and all 7 joints were actuated in the 7~DoF condition.

The 2~DoFs models of the NAO and KUKA are used to study the role of feedback frequency in 2-dimensional task spaces. In addition, The 4~DoFs and 7~DoFs conditions of NAO and KUKA are examples of more complex 3-dimensional task spaces.

\subsection{Task and Reward}
\label{sec:reward}
All experiments aim to generate controllers that can reach arbitrary goal zones in task space while controlling the robot arms in joint space.
A sparse reward function is used, i.e., reaching the goal zone leads to a reward of 1. All other actions result in a reward of 0. Such a reward function allows us to isolate the effect of feedback and analyse the learning dynamics more easily. 
Moreover, adding other rewards signals, such as punishment signals, might have a detrimental effect on learning speed \cite[e.g.,][]{Navarro-Guerrero2017Improving, Navarro-Guerrero2017Effects}, which makes both analysis and design of the reward function difficult.

The \textit{Goal Zone Radius} (GZR) for both NAO configurations is 17.5mm, and 150mm for the KUKA arm conditions.

\subsection{Interactive RL Setup}
\label{sec:irl}
\begin{figure}[htb]
    \centering
    \includegraphics[width=0.7\columnwidth]{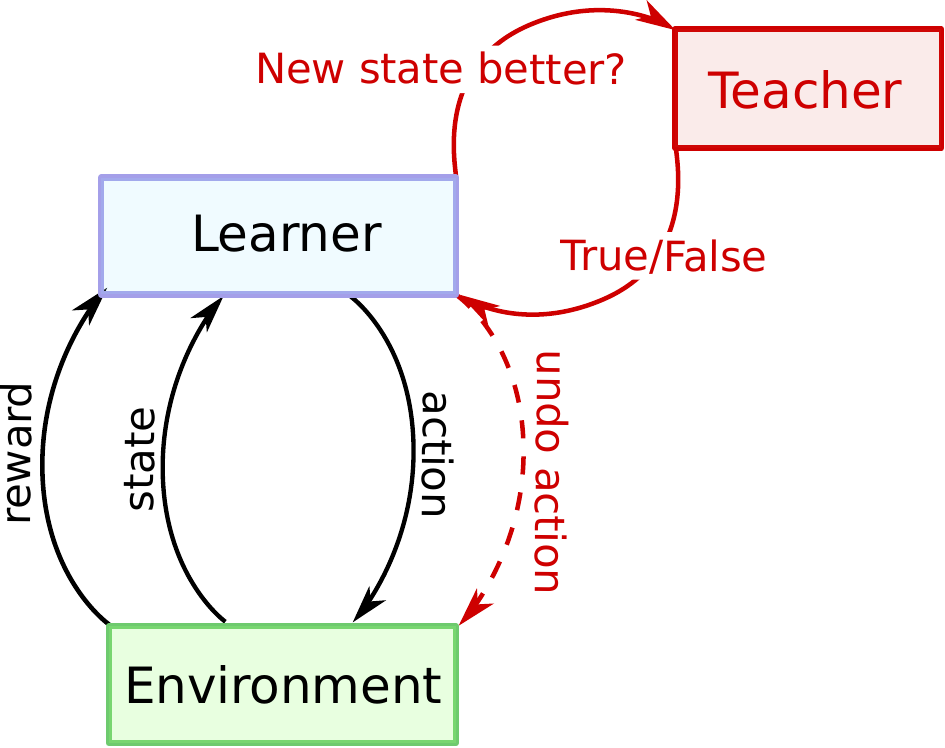}
    \caption{Diagram representing the basic RL setup, with the interactive components 
    	colored in red.}
    \label{fig:irl-our-implementation}
\end{figure}

We use the Continuous Actor-Critic Learning Automaton (CACLA) \cite{vanHasselt2007Reinforcement} as the underlying reinforcement learning framework.
The agent has an Ask Likelihood ($\mathcal{L}$) parameter, representing the likelihood of the agent asking/receiving guidance from the teacher. 
The teacher judges the agent's last action based on the Euclidean distance between the end-effector and the goal. However, the feedback to the agent is binary, and it is only used to guide the exploration process and not as an additional reward or to shape the policy directly. 
In particular, if the last action increases the distance to the goal, it is considered a mistake. When the teacher reports a mistake, the agent undoes the action and explores a new one, after which the cycle is repeated. This process is illustrated in Figure \ref{fig:irl-our-implementation}.

The stochastic feedback strategy used here is a good analogy for teachers taking breaks when providing feedback. Although this type of feedback is easy to automate, it might not always be correct. For instance, in the presence of obstacles or complex task spaces, it might be required first to move away from the goal position to reach it. 
Thus, $\mathcal{L}$ cannot be equal to 1. To prevent the agent from potentially getting stuck, we keep the maximum value of 0.99 used by Stahlhut et al.\ \cite{Stahlhut2015Interaction, Stahlhut2015Interactiona}.

\subsection{State and Action Spaces}
\label{sec:spaces}
The state space $\mathcal{S}$ for all conditions consists of the corresponding joint positions (proprioception) and the Cartesian coordinates of the target position (exteroception). 
For all conditions, the action space is limited to the maximal allowed joint displacement per time step of $\sfrac{\pi}{10}$. I.e. 
$\mathcal{A} = [- \sfrac{\pi}{10}, \sfrac{\pi}{10}]^{DoF}$.

\subsection{The Controller}
\label{sec:controllers}
The Actor and Critic are implemented with two separate multilayer perceptrons (MLPs). The networks share the same input vector. However, the networks are tuned separately using hyperparameter optimization as described in Section \ref{sec:hyperopt}. Thus, the learning rate, number of hidden layers and outputs, etc.\ may differ between the Actor and the Critic. All input and output values are scaled to the range $[-1, 1]$. The activation function for the output units is linear. The networks were implemented in PyTorch 1.10.0 and trained using Adam \cite{Kingma2015Adam}.

\subsection{Episodes}
\label{sec:episodes}
Based on both the maximum range of motion of the joints and the maximal action size, the smallest number of steps needed to traverse the entire task space was computed as follows: 
\[Steps_{min} = \frac{Range_{max}}{Action_{max}}\] 

$Steps_{min}$ was then used to define the episode length $Steps_{max}$ as $3 \times Steps_{min}$ rounded to the next tenth. 

The minimum number of goals zones $G_{min}$ to cover the entire workspace was used to determine the number of episodes $N_{train}$ per epoch. 
$G_{min}$ was calculated using optimal disc (2D task space) or ball (3D task space) packing in the task space volume. $N_{train}$ results from $10 \times G_{min}$ rounded to the second leading digit. 
Table \ref{tab:learning-settings} shows a summary of the episode parameters conditions.

\begin{table}[htbp]
    \centering
    \caption{Boundary conditions for the episodes. $Steps_{min}$: min.\ \# of steps to cover the task space, $Steps_{max}$: max.\ \# of steps the agent can take to reach the target, $G_{min}$: min.\ \# of goals to cover the task space and $N_{train}$: \# of episodes per epoch.}
    \label{tab:learning-settings}
    \begin{tabular}{@{}l@{\hspace{1.5mm}}| @{\hspace{1.9mm}}c @{\hspace{1.9mm}}c @{\hspace{1.9mm}}|c @{\hspace{1.9mm}}c @{\hspace{1.9mm}}c@{}}
      \hline\noalign{\smallskip}
        & \multicolumn{2}{c|}{NAO} & \multicolumn{3}{c}{KUKA} \\
        & 2~DoF & 4~DoF & 2~DoF & 4~DoF & 7~DoF \\
      \noalign{\smallskip}\hline\noalign{\smallskip}
      GZR & 17.5mm & 17.5mm & 150mm & 150mm & 150mm \\
      $Steps_{min}$ & 6 & 13 & 19 & 19 & 19  \\
      $Steps_{max}$ & 20 & 40 & 60 & 60 & 60 \\
      $G_{min}$ & 22 & 224 & 38 & 261 & 261  \\
      $N_{train}$ & 220 & 2200 & 380 & 2600 & 2600  \\
      \noalign{\smallskip}\hline
    \end{tabular}
\end{table}

\subsection{Performance Metrics}
\label{sec:metrics}
The following metrics were used to quantify the effect of feedback frequency. Lower values indicate better performance:
\begin{itemize}
  \item \textit{Positioning error}: mean Euclidean distance to the target divided by the target radius.
  \item \textit{Failure rate}: percentage of missed targets during evaluation, which is equivalent to \\{$\textrm{\textit{Failure rate}} = 1 - \textrm{average cumulative reward}$}.
\end{itemize}

The performance learning curves are analysed with respect to the cumulative steps instead of epochs since it better reflects the total amount of interaction with the environment. 
The slope of the failure rate is used as an indicator of the \textit{improvement speed}.

We furthermore consider thresholded performance profiles. Note that for visual clarity, the error bars represent the standard error of the mean and not a confidence interval.
Here, the steps of the $\mathcal{L}$ agent reaching the corresponding \textit{failure rate} threshold first are used as reference. 
Then the best \textit{failure rate} up to this step count is compared across different $\mathcal{L}$ values. 
For this analysis, we used a two-sided Wilcoxon rank sum test with respect to the $\mathcal{L} = 0.0$ condition. 
This temporal threshold strategy copes better when dealing with conditions that cannot achieve an arbitrary performance threshold than the \textit{time to threshold} \cite{Taylor2009Transfer} strategy.

\subsection{Datasets}
Following similar practices as in supervised learning, three datasets were used: a training, a validation, and a test set. Both the training set and test set are of the same size $N_{train}$, while the validation set is $\sfrac{1}{5}$ the size of the training set, see Table~\ref{tab:learning-settings}. 
The datasets consist of pairs of initial positions for the agent and the target. These positions are generated randomly from a uniform distribution in joint space. The targets are represented in Cartesian coordinates and result from feeding random joint configurations into the forward kinematics model of the corresponding robotic arm. Target coordinates that lie within the goal zone of the corresponding initial position are rejected. One epoch is defined as training for all pairs in the training set once.

\subsection{Hyperparameter Optimization}
\label{sec:hyperopt}
Hyperopt \cite{Bergstra2013Hyperopt} was used to determine the best hyperparameters out of 100 hyperparameter sets for each experimental condition.
Preliminary tests showed signs of significant performance improvement by the 10$^{th}$ (2~DoFs) or 20$^{th}$ (4~DoFs and 7~DoFs) epoch. 
Thus, during hyperparameter selection, the 2~DoFs conditions agents were trained for 10 epochs while the 4~DoF and 7~DoFs conditions were trained for 20 epochs. In all cases, we used the corresponding training set. 
The best hyperparameters set was selected based on the lowest \textit{positioning error} in the validation set at the respective final epoch. 

Prior tests showed that optimizing for the lowest \textit{positioning error} or fastest convergence speed leads to similar results. In real-world scenarios, it is arguably more important to have the robotic arm performing the defined task precisely -- with minimal possible error -- than learn to perform quickly but with low repeatability or precision. Thus, here the hyperparameters were optimized for the lowest \textit{positioning error}.

The hyperparameter search can be done at least in two manners: 1) optimizing the hyperparameters for each Ask Likelihood ($\mathcal{L}$) value independently, or 2) optimizing the hyperparameters only for $\mathcal{L}=0.0$ and evaluating the performance for increasing values of $\mathcal{L}$. 
The latter was selected for two reasons. Firstly, hyperparameter optimization is computationally expensive. Secondly, this strategy allows for quantifying the gain of using a particular feedback frequency in an existing system of vanilla RL ($\mathcal{L}=0.0$).

Table \ref{tab:hyperparameters} summarizes the hyperparameters and optimization boundaries. The last five hyperparameters corresponding to the neural network configuration are optimized independently for the Actor and Critic, but share the same ranges. 

\begin{table}[htbp]
    \caption{List of hyperparameters to be optimized and their interval of possible values.}
    \label{tab:hyperparameters}
    \begin{tabular}{ll}
        \hline\noalign{\smallskip}
        Hyperparameter & Range \\
        \noalign{\smallskip}\hline\noalign{\smallskip}
        Actor learning rate & [$10^{-4}, 10^{-2}$] \\
        Critic learning rate & [$10^{-4}, 10^{-2}$] \\
        Exploration rate & [0.2, 0.9] \\
        Discount factor & [0.75, 1.0] \\
        Zeta & [$10^{-4}, 10^{-1.3}$] \\
        Initial variance & [1.0, 3.0] \\
        \# of hidden layers & 1, 2 or 3 \\
        \# of neurons on 1\textsuperscript{st} layer & [10, 100] step of 10 \\
        \# of Neurons on 2\textsuperscript{nd} layer & [5, 100] step of 5 \\
        \# of Neurons on 3\textsuperscript{rd} layer & [5, 100] step of 5 \\
        Activation function & ReLu, SeLu, or Softplus\\
        \noalign{\smallskip}\hline
    \end{tabular}
\end{table}

\subsection{Training and Testing}
The agents are trained on the same training set used for the hyperparameter optimization, but this time the agents' performance is evaluated on the test set. 
All agents are trained for a fixed number of epochs. 
Eleven agent versions for each condition were trained, including the baseline agents for $\mathcal{L}=0.0$ (non-interactive) and ten other agents sets with $\mathcal{L}$ values increasing in increments of 0.1, with the last agent having $\mathcal{L}=0.99$ (fully interactive). 
The $\mathcal{L}$ values do not change during training. 
During testing, no learning occurs, no interaction is possible, and no undo action is performed. 
Statistics are taken over 20 randomly initialized agents for each task and each $\mathcal{L}$ value.

\section{Results}
\label{sec:results}
In the following section, we present the temporal evolution of the \textit{failure rate} on the test sets for all experimental conditions.
The failure rate value at each point is the average over the 20 randomly initialized agents.
 
In addition, for each experimental condition, we compare the \textit{failure rate} for different $\mathcal{L}$ values at various threshold levels to assess the optimal $\mathcal{L}$ value at different stages in training.

\subsection{NAO 2~DoFs Experiment}
\begin{figure}[!htb]
    	\centering
    	\subfloat[][]{\includegraphics[width=\columnwidth]{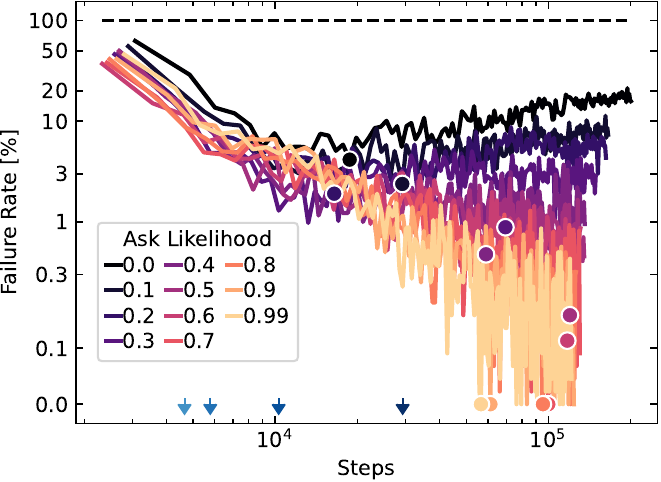}
    	\label{fig:evolution-nao2}}
    	\\*[-1mm]
    	\subfloat[][]{\includegraphics[width=\columnwidth]{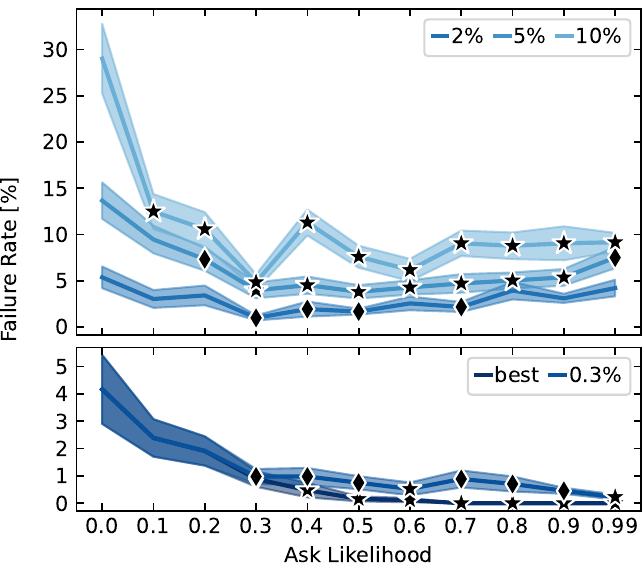}
    	\label{fig:nao2-failure-rates}}
	    \caption{\textbf{(a)}: \textit{Failure rate} evolution for the NAO 2~DoFs experiment in log scale. The circle indicates the best performance for the corresponding $\mathcal{L}$ value for the whole training. The blue arrows show the number of environment steps needed for the fastest $\mathcal{L}$ agents to reach $10\%$, $5\%$, $2\%$, and $0.3\%$ \textit{failure rate}. \linebreak 
    	\textbf{(b)}: \textit{Time thresholded failure rates} for the NAO 2~DoFs condition. Statistical significance with respect to the $\mathcal{L}=0.0$ condition was computed using a two-sided Wilcoxon rank sum test. The \ding{117} show significance with respect to $\mathcal{L} = 0.0$ at $p<0.05$, while the \ding{72} show significance with respect to $\mathcal{L} = 0.0$ at $p<0.001$.}
\end{figure}

Figure \ref{fig:evolution-nao2} shows the performance evolution for the NAO 2~DoFs condition. The circles indicate the best \textit{failure rate} value achieved during training for each $\mathcal{L}$ value. 
The \textit{failure rate} continuously improves with higher $\mathcal{L}$ values, even reaching perfect performance in late training for $\mathcal{L} \geq 0.7$, whereas the performance for low $\mathcal{L}$ starts to diverge around $10^4$ steps. The \textit{failure rate} curves' \textit{rate of improvement} is comparable for all $\mathcal{L}$ values during the first few epochs.

Figure \ref{fig:nao2-failure-rates} shows the thresholded \textit{failure rate} performances in the NAO 2~DoFs condition. The cumulative step counts corresponding to the thresholds are marked by blue arrows in Figure \ref{fig:evolution-nao2}.
At the first threshold levels, there is a trend toward a significant lower \textit{failure rate} as $\mathcal{L}$ values increase. However, the behaviour is dynamic, and no single $\mathcal{L}$ value remains the best through training. 
Within the tested time horizon, the final performance favours the highest $\mathcal{L}$ values.

\subsection{NAO 4~DoFs Experiment}
\begin{figure}[!htb]
	\centering
	\subfloat[][]{\includegraphics[width=\columnwidth]{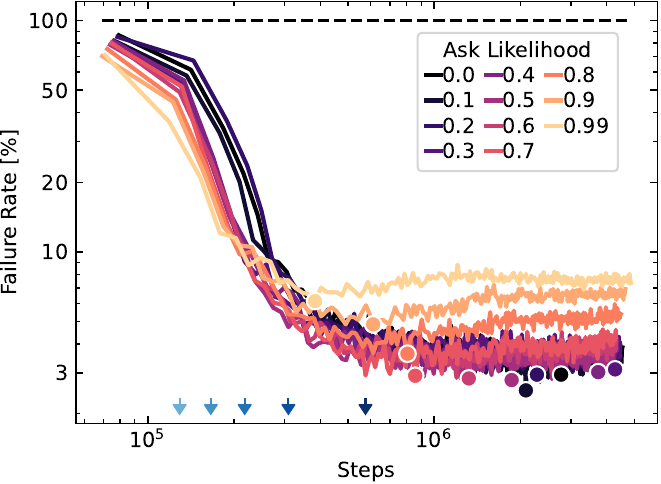} \label{fig:evolution-nao4}}
	\\*[-1mm]
	\subfloat[][]{\includegraphics[width=\columnwidth]{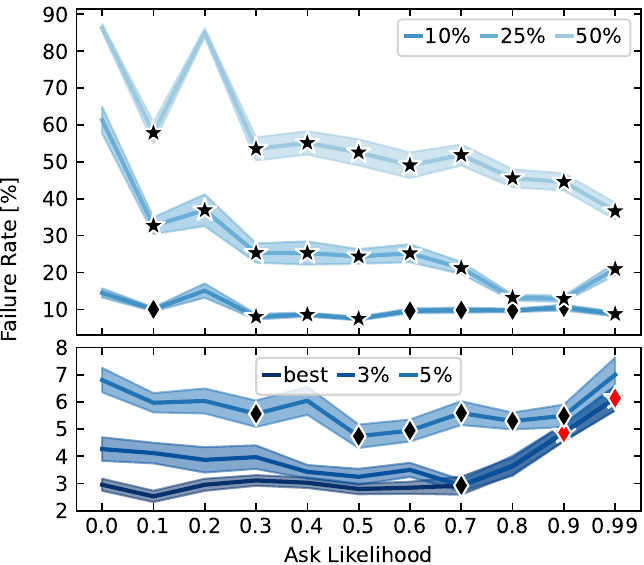}
	\label{fig:nao4-failure-rates}}
	\caption{\textbf{(a)}: \textit{Failure rate} evolution for the NAO 4~DoFs experiment shown in log scale. The circle indicates the best performance for the corresponding $\mathcal{L}$ value. The blue arrows show the number of environment steps needed for the fastest $\mathcal{L}$ agents to reach $50\%$, $25\%$ $10\%$, $5\%$, and $3\%$ \textit{failure rate}. \linebreak 
	\textbf{(b)}: \textit{Time thresholded Failure rates} for NAO 4~DoFs. Statistical significance was computed using a ranksum test. The \ding{117} show significance with respect to $\mathcal{L} = 0.0$ at $p<0.05$, while the \ding{72} show significance with respect to $\mathcal{L} = 0.0$ at $p<0.001$. Red markers indicate significantly detrimental effects.}
\end{figure}

Figure \ref{fig:evolution-nao4} shows the performance evolution for NAO 4~DoFs condition. 
Values up to $\mathcal{L} = 0.7$ convergence to a similar value. In contrast, the best performance of higher $\mathcal{L}$ is reached earlier, after which the \textit{failure rates} start to diverge slowly. 

The \textit{improvement speed} is initially faster for higher $\mathcal{L}$ values before they start to diverge.

Figure \ref{fig:nao4-failure-rates} shows the corresponding time thresholded \textit{failure rate} analysis. 
Here the highest effect on the \textit{failure rates} is observed in the first half of training at very high $\mathcal{L}$ values. 
In the later phase of training, at the $5\%$ threshold, the optimal $\mathcal{L}$ shifts towards medium and high $\mathcal{L}$ values. 
Finally, a significant benefit is mostly absent for the strictest threshold of $3\%$ and beyond. However, for $\mathcal{L} \geq 0.9$, the effect is significantly detrimental, as indicated by the red markers in Figure \ref{fig:nao4-failure-rates}.

\subsection{KUKA 2~DoFs Experiment}
\begin{figure}[!htb]
    \centering
    \subfloat[][]{\includegraphics[width=\columnwidth]{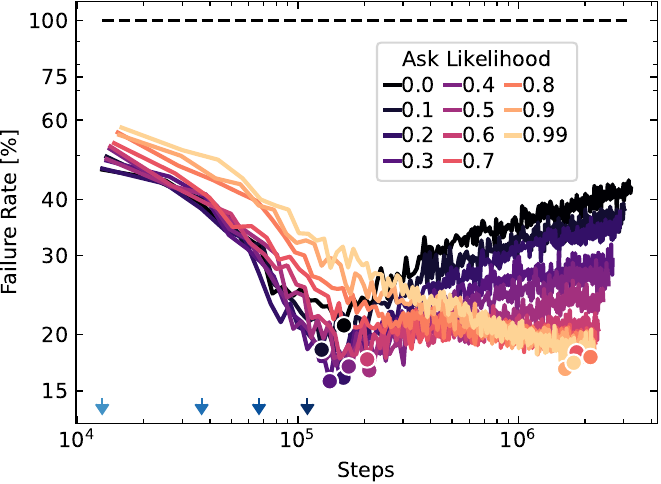}
    \label{fig:evolution-kuka2}}
	\\*[-1mm]
	\subfloat[][]{\includegraphics[width=\columnwidth]{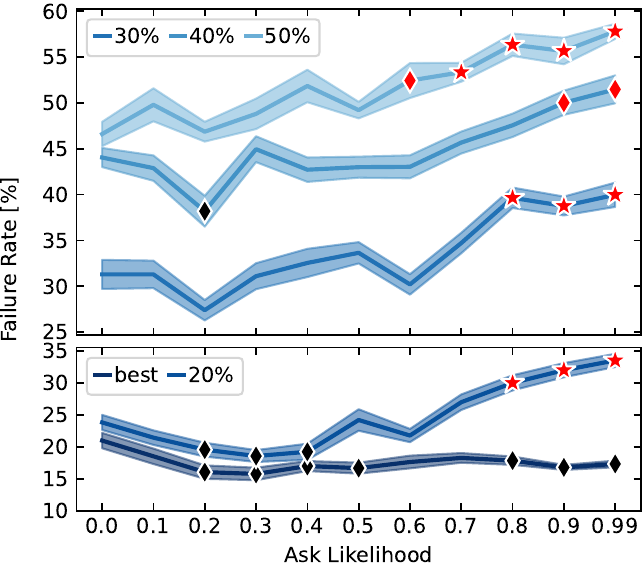}
	\label{fig:kuka2-failure-rates}}
    \caption{\textbf{(a)} \textit{Failure rate} evolution for the KUKA 2~DoFs experiment shown in log scale. The circle indicates the best performance for the corresponding $\mathcal{L}$ value. The blue arrows show the number of environment steps needed for the fastest $\mathcal{L}$ agents to reach $50\%$, $40\%$, $30\%$, and $20\%$ \textit{failure rate}. \linebreak
    \textbf{(b)} \textit{Time thresholded failure rates} for the KUKA 2~DoFs condition. Statistical significance was computed using a two-sided Wilcoxon rank sum test. The \ding{117} show significance with respect to $\mathcal{L} = 0.0$ at $p<0.05$, while the \ding{72} show significance with respect to $\mathcal{L} = 0.0$ at $p<0.001$. Red markers indicate significantly detrimental effects.}
\end{figure}

Figure \ref{fig:evolution-kuka2} shows the performance evolution for the KUKA 2~DoFs condition. The overall \textit{failure rate} in this condition is relatively high. However, a similar trend to that of the NAO 2~DoF condition can be observed, i.e., the \textit{failure rate} continuously improves with a higher $\mathcal{L}$ value. In contrast, the performance for low $\mathcal{L}$ values starts to diverge after the respective best performance is reached.

The \textit{improvement speed} for low to medium $\mathcal{L}$ agents is initially higher. 
Whereas the \textit{improvement speed} for higher $\mathcal{L}$ values is slightly lower, it is maintained almost constantly throughout the tested time horizon.

Figure \ref{fig:kuka2-failure-rates} shows the time thresholded \textit{failure rate} for the KUKA 2~DoFs condition. Unlike in both NAO conditions, here, early in training, the interaction does not yield any benefits. Interaction is even significantly detrimental for very high $\mathcal{L}$ values. 
Only towards the end of training does interaction significantly reduce the \textit{failure rate}.

\subsection{KUKA 4~DoFs Experiment}
\begin{figure}[!tb]
    \centering
    \subfloat[][]{\includegraphics[width=\columnwidth]{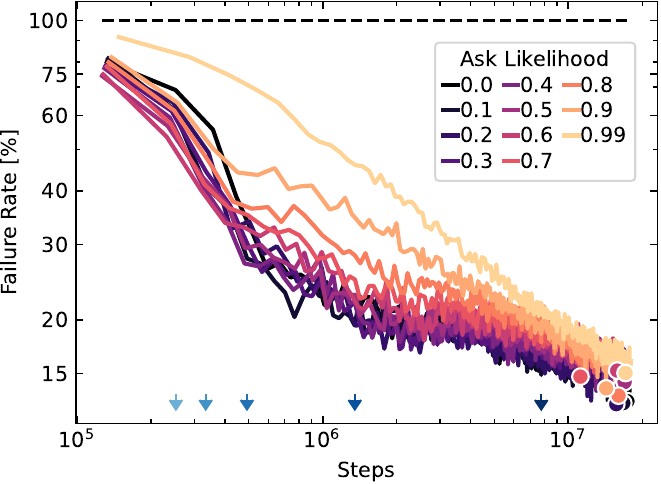}
    \label{fig:evolution-kuka4}}
	\\*[-1mm]
	\subfloat[][]{\includegraphics[width=\columnwidth]{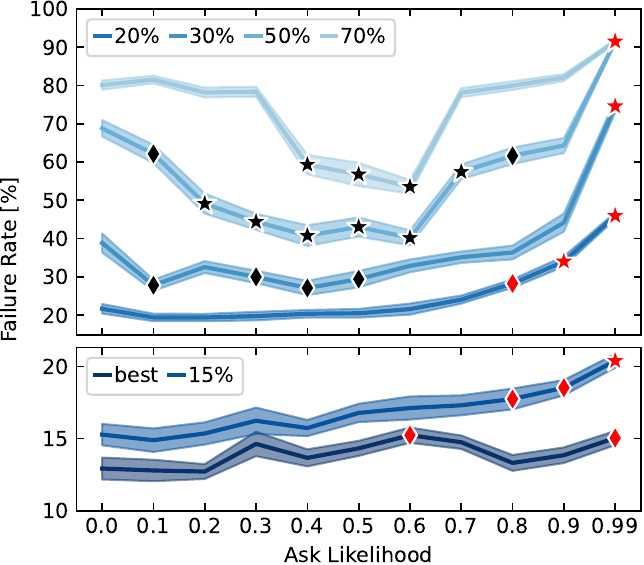}
	\label{fig:kuka4-failure-rates}}
    \caption{\textbf{(a)} \textit{Failure rate} evolution for the KUKA 4~DoFs experiment shown in log scale. The circle indicates the best performance for the corresponding $\mathcal{L}$ value. The blue arrows show the number of environment steps needed for the fastest $\mathcal{L}$ agents to reach $70\%$, $50\%$, $30\%$, $20\%$, and $15\%$ \textit{failure rate}. \linebreak
    \textbf{(b)} \textit{Time thresholded failure rates} for the KUKA 4~DoFs condition. Statistical significance was computed using a two-sided Wilcoxon rank sum test. The \ding{117} show significance with respect to $\mathcal{L} = 0.0$ at $p<0.05$, while the \ding{72} show significance with respect to $\mathcal{L} = 0.0$ at $p<0.001$. Red markers indicate a significantly detrimental effects.}
\end{figure}

Figure \ref{fig:evolution-kuka4} shows the performance evolution for the KUKA 4~DoFs condition. 
Again, the overall \textit{failure rate} in this condition is relatively high.
In contrast to all other experimental conditions, all $\mathcal{L}$ values lead to a monotonically improving \textit{failure rate}. Lower $\mathcal{L}$ values initially show a faster rate of improvement but slow down as learning progresses. In contrast, very high $\mathcal{L}$ values display a lower rate of improvement, which is, however, almost constant throughout the tested time horizon.

Figure \ref{fig:kuka4-failure-rates} shows the time thresholded \textit{failure rate} for the KUKA 4~DoFs condition. 
Early in training, interaction leads to a significant reduction in \textit{failure rate} primarily for intermediate $\mathcal{L}$ values. In contrast, the highest $\mathcal{L}$ values have a significantly detrimental effect on performance throughout the tested time horizon. However, from the data, it cannot be judged what the very long-term behaviour of the agents will be since the performance has not converged.

\subsection{KUKA 7~DoFs Experiment}
\begin{figure}[!htb]
    \centering
    \subfloat[][]{\includegraphics[width=\columnwidth]{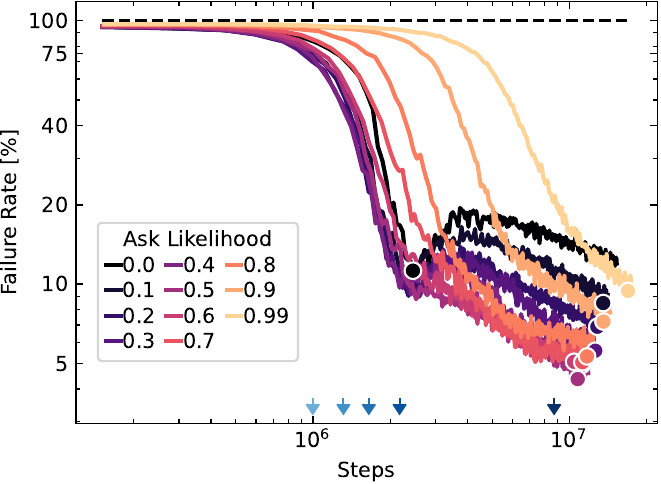}
    \label{fig:evolution-kuka7}}
	\\*[-1mm]
	\subfloat[][]{\includegraphics[width=\columnwidth]{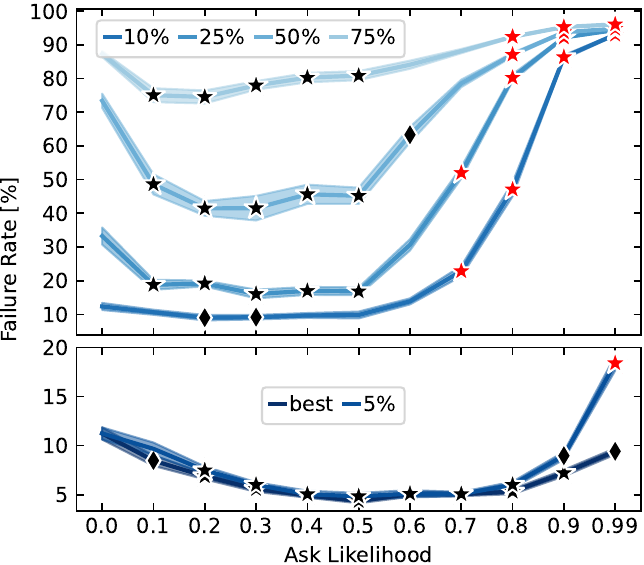}
	\label{fig:kuka7-failure-rates}}
    \caption{\textbf{(a)} \textit{Failure rate} evolution for the KUKA 7~DoFs experiment shown in log scale. The circle indicates the best performance for the corresponding $\mathcal{L}$ value. The blue arrows show the number of environment steps needed for the fastest $\mathcal{L}$ agents to reach $75\%$, $50\%$, $25\%$, $10\%$ and $5\%$ \textit{failure rate}. \linebreak
    \textbf{(b)} \textit{Time thresholded failure rates} for the KUKA 7~DoFs condition. Statistical significance was computed using a two-sided Wilcoxon rank sum test. The \ding{117} show significance with respect to $\mathcal{L} = 0.0$ at $p<0.05$, while the \ding{72} show significance with respect to $\mathcal{L} = 0.0$ at $p<0.001$. Red markers indicate a significantly detrimental effects.}
\end{figure}

Figure \ref{fig:evolution-kuka7} shows the performance evolution for the KUKA 7~DoFs condition. 
As for the 2~DoFs conditions, low to medium $\mathcal{L}$ agents reach a local optimum around $2\times10^6$ steps, after which the performance temporarily deteriorates. 
However, in contrast to the 2~DoFs conditions, the performance continues to improve beyond the initial local optimum. 
Higher $\mathcal{L}$ agents have a lower rate of improvement but do not experience any divergent behaviour, at least within the tested time horizon.

Figure \ref{fig:kuka7-failure-rates} shows the time thresholded \textit{failure rate} for the KUKA 7~DoFs condition. 
Similarly, as in the previous KUKA conditions, low to medium $\mathcal{L}$ values lead to significantly better performance early in training. In contrast, very high $\mathcal{L}$ values lead to significantly worse \textit{failure rates}. The detrimental effect becomes stronger the higher the $\mathcal{L}$ value.
Only for longer training horizons do high $\mathcal{L}$ values start to become significantly beneficial, albeit not optimal. The long-term behaviour cannot be clearly judged here since the performance has not converged.

Table \ref{tab:learning-stats} shows a combined summary of the statistical analyses of the effects on the \textit{failure rate} thresholds for all tested robotic tasks and $\mathcal{L}$ values. The effect size reported is the difference of means. 

\begin{table*}[!htbp]
	\centering
    \caption{Effect Size for different values of $\mathcal{L}$ across robot tasks, computed as difference of means. Statistical significance was determined by a Wilcoxon ranksum test. The \ding{117} show significance with respect to $\mathcal{L} = 0.0$ at $p<0.05$, while the \ding{72} show significance with respect to $\mathcal{L} = 0.0$ at $p<0.001$. Red markers indicate significantly 
    detrimental effects.}
    \label{tab:learning-stats}
    \fontsize{8bp}{9.5bp}\selectfont
	\begin{tabular}{@{\hspace{-8pt}}c@{}r|rrrrrrrrrr@{}}
 \multicolumn{1}{l}{}              & \multicolumn{1}{c|}{Failure } & \multicolumn{10}{c}{Ask Likelihood}                                                                                                                                                                                                                                 \\
    & \multicolumn{1}{c|}{Rate}   & \multicolumn{1}{c}{0.1} & \multicolumn{1}{c}{0.2} & \multicolumn{1}{c}{0.3} & \multicolumn{1}{c}{0.4} & \multicolumn{1}{c}{0.5} & \multicolumn{1}{c}{0.6} & \multicolumn{1}{c}{0.7} & \multicolumn{1}{c}{0.8} & \multicolumn{1}{c}{0.9} & \multicolumn{1}{c}{0.99}  \\ \hline

 \multirow{5}{*}{\rotcell{\hspace*{-9pt}NAO \hspace*{-9pt}2~DoFs}}
 & 10 & 
-16.55\textsuperscript{\ding{72}}&-18.45\textsuperscript{\ding{72}}&-24.18\textsuperscript{\ding{72}}&-17.73\textsuperscript{\ding{72}}&-21.43\textsuperscript{\ding{72}}&-22.86\textsuperscript{\ding{72}}&-19.98\textsuperscript{\ding{72}}&-20.25\textsuperscript{\ding{72}}&-19.98\textsuperscript{\ding{72}}&-19.86\textsuperscript{\ding{72}}
 \\

 & 5 & 
-&-6.34\textsuperscript{\ding{117}}&-9.7\textsuperscript{\ding{72}}&-9.14\textsuperscript{\ding{72}}&-9.86\textsuperscript{\ding{72}}&-9.39\textsuperscript{\ding{72}}&-8.95\textsuperscript{\ding{72}}&-8.64\textsuperscript{\ding{72}}&-8.3\textsuperscript{\ding{72}}&-6.16\textsuperscript{\ding{117}}
 \\

 & 2 &
-&-&-4.39\textsuperscript{\ding{117}}&-3.43\textsuperscript{\ding{117}}&-3.7\textsuperscript{\ding{117}}&-&-3.2\textsuperscript{\ding{117}}&-&-&-
 \\

 & 0.3 &
-&-&-3.18\textsuperscript{\ding{117}}&-3.18\textsuperscript{\ding{117}}&-3.41\textsuperscript{\ding{117}}&-3.64\textsuperscript{\ding{72}}&-3.27\textsuperscript{\ding{117}}&-3.45\textsuperscript{\ding{117}}&-3.7\textsuperscript{\ding{117}}&-3.93\textsuperscript{\ding{72}}
  \\

 & best &
-&-&-3.27\textsuperscript{\ding{117}}&-3.68\textsuperscript{\ding{72}}&-4.0\textsuperscript{\ding{72}}&-4.05\textsuperscript{\ding{72}}&-4.16\textsuperscript{\ding{72}}&-4.16\textsuperscript{\ding{72}}&-4.16\textsuperscript{\ding{72}}&-4.16\textsuperscript{\ding{72}}
  \\ \hline

 \multirow{6}{*}{\rotcell{\hspace*{-9pt}NAO \hspace*{-9pt}4~DoFs}}  
 & 50 &
-28.51\textsuperscript{\ding{72}}&-&-32.84\textsuperscript{\ding{72}}&-31.23\textsuperscript{\ding{72}}&-33.8\textsuperscript{\ding{72}}&-37.29\textsuperscript{\ding{72}}&-34.5\textsuperscript{\ding{72}}&-40.82\textsuperscript{\ding{72}}&-41.78\textsuperscript{\ding{72}}&-49.75\textsuperscript{\ding{72}}
 \\

 & 25 &
-28.61\textsuperscript{\ding{72}}&-24.42\textsuperscript{\ding{72}}&-36.0\textsuperscript{\ding{72}}&-36.01\textsuperscript{\ding{72}}&-36.94\textsuperscript{\ding{72}}&-36.09\textsuperscript{\ding{72}}&-40.06\textsuperscript{\ding{72}}&-48.1\textsuperscript{\ding{72}}&-48.43\textsuperscript{\ding{72}}&-40.36\textsuperscript{\ding{72}}
 \\

 & 10 &
-4.46\textsuperscript{\ding{117}}&-&-6.45\textsuperscript{\ding{72}}&-5.87\textsuperscript{\ding{72}}&-6.96\textsuperscript{\ding{72}}&-4.86\textsuperscript{\ding{117}}&-4.65\textsuperscript{\ding{117}}&-4.75\textsuperscript{\ding{117}}&-3.97\textsuperscript{\ding{117}}&-5.71\textsuperscript{\ding{72}}
 \\

 & 5 &
-&-&-1.24\textsuperscript{\ding{117}}&-&-2.07\textsuperscript{\ding{117}}&-1.86\textsuperscript{\ding{117}}&-1.21\textsuperscript{\ding{117}}&-1.51\textsuperscript{\ding{117}}&-1.31\textsuperscript{\ding{117}}&-
 \\

 & 3 & 
-&-&-&-&-&-&-1.34\textsuperscript{\ding{117}}&-&{\color{red} 0.6\textsuperscript{\ding{117}}}&{\color{red} 1.88\textsuperscript{\ding{117}}}
 \\

 & best & 
-&-&-&-&-&-&-&-&{\color{red} 1.92\textsuperscript{\ding{72}}}&{\color{red} 3.19\textsuperscript{\ding{72}}}
 \\ \hline

 \multirow{5}{*}{\rotcell{\hspace*{-9pt}KUKA \hspace*{-9pt}2~DoFs}} 
 & 50 & 
-&-&-&-&-&{\color{red} 5.83\textsuperscript{\ding{117}}}&{\color{red} 6.74\textsuperscript{\ding{72}}}&{\color{red} 9.74\textsuperscript{\ding{72}}}&{\color{red} 9.05\textsuperscript{\ding{72}}}&{\color{red} 11.2\textsuperscript{\ding{72}}}
 \\

 & 40 & 
-&-5.86\textsuperscript{\ding{117}}&-&-&-&-&-&-&{\color{red} 5.97\textsuperscript{\ding{117}}}&{\color{red} 7.43\textsuperscript{\ding{117}}}
 \\

 & 30 &
-&-&-&-&-&-&-&{\color{red} 8.36\textsuperscript{\ding{72}}}&{\color{red} 7.46\textsuperscript{\ding{72}}}&{\color{red} 8.68\textsuperscript{\ding{72}}}
 \\

 & 20 &
-&-4.26\textsuperscript{\ding{117}}&-5.26\textsuperscript{\ding{117}}&-4.59\textsuperscript{\ding{117}}&-&-&-&{\color{red} 6.2\textsuperscript{\ding{72}}}&{\color{red} 8.18\textsuperscript{\ding{72}}}&{\color{red} 9.71\textsuperscript{\ding{72}}}
 \\

 & best & 
-&-4.93\textsuperscript{\ding{117}}&-5.22\textsuperscript{\ding{117}}&-3.99\textsuperscript{\ding{117}}&-4.32\textsuperscript{\ding{117}}&-&-&-3.14\textsuperscript{\ding{117}}&-4.2\textsuperscript{\ding{117}}&-3.66\textsuperscript{\ding{117}}
  \\ \hline

 \multirow{6}{*}{\rotcell{\hspace*{-9pt}KUKA \hspace*{-9pt}4~DoFs}}
 & 70 & 
-&-&-&-20.78\textsuperscript{\ding{72}}&-23.34\textsuperscript{\ding{72}}&-26.56\textsuperscript{\ding{72}}&-&-&-&{\color{red} 11.4\textsuperscript{\ding{72}}}
 \\

 & 50 & 
-6.68\textsuperscript{\ding{117}}&-19.67\textsuperscript{\ding{72}}&-24.42\textsuperscript{\ding{72}}&-28.04\textsuperscript{\ding{72}}&-25.79\textsuperscript{\ding{72}}&-28.63\textsuperscript{\ding{72}}&-11.38\textsuperscript{\ding{72}}&-7.19\textsuperscript{\ding{117}}&-&{\color{red} 22.68\textsuperscript{\ding{72}}}
 \\

 & 30 &
-11.05\textsuperscript{\ding{117}}&-&-8.87\textsuperscript{\ding{117}}&-11.76\textsuperscript{\ding{117}}&-9.46\textsuperscript{\ding{117}}&-&-&-&-&{\color{red} 35.71\textsuperscript{\ding{72}}}
 \\

 & 20 &
-&-&-&-&-&-&-&{\color{red} 6.6\textsuperscript{\ding{117}}}&{\color{red} 12.35\textsuperscript{\ding{72}}}&{\color{red} 24.3\textsuperscript{\ding{72}}}
 \\

 & 15 &
-&-&-&-&-&-&-&{\color{red} 2.47\textsuperscript{\ding{117}}}&{\color{red} 3.25\textsuperscript{\ding{117}}}&{\color{red} 5.12\textsuperscript{\ding{72}}}
 \\
 
 & best & 
-&-&-&-&-&{\color{red} 2.31\textsuperscript{\ding{117}}}&-&-&-&{\color{red} 2.11\textsuperscript{\ding{117}}}
  \\ \hline

 \multirow{6}{*}{\rotcell{\hspace*{-9pt}KUKA \hspace*{-9pt}7~DoFs}}
 & 75 & 
-12.23\textsuperscript{\ding{72}}&-12.76\textsuperscript{\ding{72}}&-9.33\textsuperscript{\ding{72}}&-7.02\textsuperscript{\ding{72}}&-6.42\textsuperscript{\ding{72}}&-&-&{\color{red} 5.12\textsuperscript{\ding{72}}}&{\color{red} 8.01\textsuperscript{\ding{72}}}&{\color{red} 8.8\textsuperscript{\ding{72}}}
 \\

 & 50 & 
-24.79\textsuperscript{\ding{72}}&-31.95\textsuperscript{\ding{72}}&-31.89\textsuperscript{\ding{72}}&-27.76\textsuperscript{\ding{72}}&-28.21\textsuperscript{\ding{72}}&-10.08\textsuperscript{\ding{117}}&-&{\color{red} 13.71\textsuperscript{\ding{72}}}&{\color{red} 20.35\textsuperscript{\ding{72}}}&{\color{red} 21.39\textsuperscript{\ding{72}}}
 \\

 & 25 &
-14.47\textsuperscript{\ding{72}}&-14.08\textsuperscript{\ding{72}}&-17.21\textsuperscript{\ding{72}}&-16.28\textsuperscript{\ding{72}}&-16.41\textsuperscript{\ding{72}}&-&{\color{red} 18.76\textsuperscript{\ding{72}}}&{\color{red} 47.02\textsuperscript{\ding{72}}}&{\color{red} 58.95\textsuperscript{\ding{72}}}&{\color{red} 61.16\textsuperscript{\ding{72}}}
 \\

 & 10 &
-&-3.33\textsuperscript{\ding{117}}&-3.18\textsuperscript{\ding{117}}&-&-&-&{\color{red} 10.48\textsuperscript{\ding{72}}}&{\color{red} 34.77\textsuperscript{\ding{72}}}&{\color{red} 74.01\textsuperscript{\ding{72}}}&{\color{red} 80.7\textsuperscript{\ding{72}}}
 \\

 & 5 & 
-&-3.76\textsuperscript{\ding{72}}&-5.23\textsuperscript{\ding{72}}&-6.19\textsuperscript{\ding{72}}&-6.4\textsuperscript{\ding{72}}&-6.14\textsuperscript{\ding{72}}&-6.18\textsuperscript{\ding{72}}&-5.23\textsuperscript{\ding{72}}&-2.27\textsuperscript{\ding{117}}&{\color{red} 7.16\textsuperscript{\ding{72}}}
 \\

 & best &
-2.77\textsuperscript{\ding{117}}&-4.35\textsuperscript{\ding{72}}&-5.66\textsuperscript{\ding{72}}&-6.3\textsuperscript{\ding{72}}&-6.88\textsuperscript{\ding{72}}&-6.17\textsuperscript{\ding{72}}&-6.19\textsuperscript{\ding{72}}&-5.93\textsuperscript{\ding{72}}&-4.04\textsuperscript{\ding{72}}&-1.82\textsuperscript{\ding{117}}
	\end{tabular}
\end{table*}

\section{Discussion}
\label{sec:discussion}
Our thorough investigation of the Ask Likelihood's effect on the evolution of the \textit{failure rate} over time allows us to make more nuanced statements on task-dependent effects of the interaction rate than previously reported. Furthermore, our experiments can unify seemingly contradictory statements on the best choice of $\mathcal{L}$.
In summary, across the different experimental conditions, we make three main observations: 1) \textit{policy robustness}, 2) \textit{optimal 
long-term $\mathcal{L}$}, and 3) \textit{optimal $\mathcal{L}$ trajectory}. 

\paragraph{1) Policy robustness:}
\emph{With the exception of the KUKA 4~DoF condition, low $\mathcal{L}$ agents are prone to suffer from performance divergence after reaching an initial local optimum. High $\mathcal{L}$ agents in the same condition do not show this behaviour---see the KUKA 7~DoF, NAO 2~DoF, and KUKA 2~DoF conditions.}

The divergence could be caused by the limited time horizon of the hyperparameter optimization. 
In all cases where divergence occurs, it sets in after the number of epochs used for the optimization. In this case, high $\mathcal{L}$ agents not suffering from divergence would be in line with Stahlhut et al.\ \cite{Stahlhut2015Interaction, Stahlhut2015Interactiona}, who report that high $\mathcal{L}$ values lead to more robust policies and are less sensitive to optimally tuned hyperparameters.

However, we note that the seemingly fast divergence is in part attributed to the log-log scale of the figures: the number of steps at which the agents stay close to their local optimum is relatively large compared to the initially fast convergence. 
An alternative explanation is that this divergence happens regularly but is rarely observed or reported since the training is stopped automatically when initial convergence is reached (early stopping). 
The reason for this divergence could be the phenomenon called \textit{capacity loss}, which was only recently described. 
Lyle et al.\ \cite{Lyle2022Understanding} show that agents can lose the ability to adjust their value function approximator in light of new prediction errors. Capacity loss is attributed to a state representation collapse in the function approximators.
This collapse seems most prevalent in temporal difference learning algorithms, using neural networks as function approximators, and sparse and non-stationary rewards. 

An exciting question is whether high $\mathcal{L}$ agents are more robust to capacity loss. However, this investigation is beyond the scope of this paper.

\paragraph{2) Optimal long-term $\mathcal{L}$:}
\emph{High $\mathcal{L}$ values are mostly beneficial in the long run, except for the NAO 4~DoF condition. In all other conditions, high $\mathcal{L}$ agents reach either the best or comparable to the best performance observed for other $\mathcal{L}$ values, albeit at later stages in training.}

This observation agrees with Stahlhut et al.\ \cite{Stahlhut2015Interaction, Stahlhut2015Interactiona}, who report increasing performance with increasing interaction frequency. However, in the NAO 2~DoF case, there is only a small benefit of the highest $\mathcal{L}$ value over the others in the long run. Thus, the gain can be considered not very large, in accordance with Bignold et al.\ \cite{Bignold2021Evaluation}, who argue that the increased effort of very high frequent interaction does not justify the small performance gains.

\paragraph{3) Optimal $\mathcal{L}$ trajectory:}
\emph{During training within one experimental condition, the best choice of $\mathcal{L}$ depends on the proficiency level and typically changes over time. For instance, in the NAO 4~DoF experiment, the optimal $\mathcal{L}$ changes from intermediate values in early training to low values in late training. However, this pattern does not generalize across tasks. E.g., the optimal $\mathcal{L}$ changes from intermediate to high values in the NAO 2~DoF condition, in contrast to the NAO 4~DoF experiment.}

This observation can encompass the following findings:
\begin{itemize}
  \item early feedback is beneficial \cite{Cruz2016Training}, as seen in all but the KUKA 2~DoF condition, 
  \item intermediate feedback frequencies are optimal \cite{Millan-Arias2020Robust}, as seen in the NAO 2~DoF and KUKA 7~DoF conditions across most of the training,  
  \item and that too much early feedback leads to a delayed onset of learning \cite{Millan-Arias2020Robust}, as seen in all 3 KUKA conditions.
\end{itemize}

Furthermore, the shift of optimal $\mathcal{L}$ values during training leads us also to conclude that the interaction rate should be changed adaptively for optimal performance, as also suggested by Cruz et al.\ \cite{Cruz2016Training} and Stahlhut et al.\ \cite{Stahlhut2015Interaction, Stahlhut2015Interactiona}.

\begin{figure}[!tb]
	\centering
	\includegraphics[width=\columnwidth]{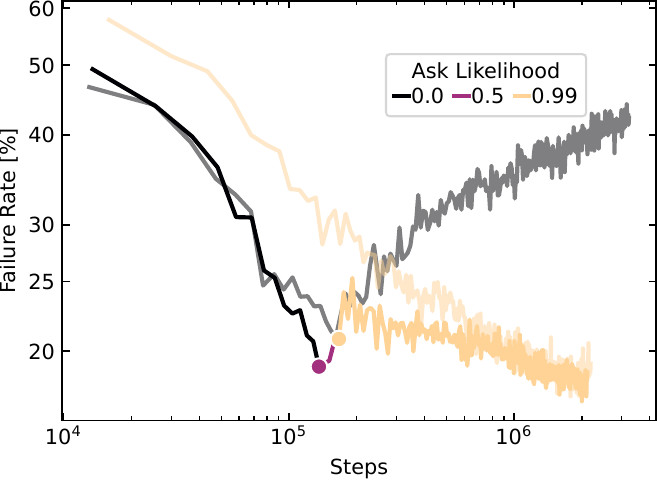}
	\caption{Adaptive \textit{failure rate} in the KUKA 2~DoF task. The initial $\mathcal{L}=0.0$ is changed to $\mathcal{L}=0.5$ at epoch 14 and $\mathcal{L}=0.99$ at epoch 18. The 50\% opacity black and yellow curves show the original performance for fixed $\mathcal{L}=0.0$ and $\mathcal{L}=0.99$, compare Figure \ref{fig:evolution-kuka2}. The adaptive agents show both features of fast early training and long-term convergence of the fixed $\mathcal{L}$ agents.}
	\label{fig:adaptive-al-kuka2}
\end{figure}

As a proof of concept, we trained agents on the KUKA 2~DoF task, starting with $\mathcal{L}=0.0$, switching to $\mathcal{L}=0.5$ at epoch 14, and finally to $\mathcal{L}=0.99$ at epoch 18. Figure~\ref{fig:adaptive-al-kuka2} shows that it is possible to achieve the early convergence of the low $\mathcal{L}$ agents in this task, combined with the long-term refinement of high $\mathcal{L}$ agents. The switch of $\mathcal{L}$ induces a short-term performance deterioration. However, measuring the performance as the area under the curve, the adaptive strategy is superior to the fixed $\mathcal{L}$ agents.

An important question is how to choose the optimal adaptive $\mathcal{L}$-strategy without having to train agents for various $\mathcal{L}$-values before. Whereas it seems to be a good rule of thumb to switch to high $\mathcal{L}$ in late training, the optimal values in early training are very diverse. Here, \textit{low} $\mathcal{L}$ values are optimal for the KUKA 2~DoF condition, \textit{intermediate} for KUKA 4~DoF, KUKA 7~DoF, and NAO 2~DoF, or \textit{high values} for the NAO 4~DoF, spanning 
a range of $\mathcal{L}=0.2$ to $\mathcal{L}=0.8$ across experiments. 

We hypothesize that the optimal $\mathcal{L}$ in early training is influenced by how much the teacher simplifies the task. Expressly, if the task for high $\mathcal{L}$ values becomes too easy in comparison to the $\mathcal{L}=0.0$ task, the agent might fail to generalize and explore too little. Note that the agent effectively faces the $\mathcal{L}=0.0$ situation during evaluation since it is not receiving feedback then. 
Specifically, consider a fully interactive agent. Here, the teacher will rarely allow actions that increase the distance of the end effector to the goal. 
This scenario simplifies the task during training but also entails that sub-optimal state-action pairs are seldomly encountered. Note that this situation primarily applies to mistake-correcting teachers as used in this article.

We quantify the reduction in task complexity by $\mathcal{L}$ as the average \textit{failure rate} of a newly initialized agent on the training set with an interaction frequency $\mathcal{L}$, but without policy updates. 
All failure rates for $\mathcal{L} \neq 0$ are normalized to the performance of the baseline $\mathcal{L} = 0$ agent. 
Each value is averaged over 20 randomly initialized agents. 
We compare the relative task complexities with the best choices of $\mathcal{L}$ at similar stages in early training. For this, we use the \textit{failure rate} thresholds of 50\% for KUKA 7~DoFs and KUKA 4~DoFs, 40\% for KUKA 2~DoFs, 25\% for NAO 4~DoFs, and 5\% for NAO 2~DoFs (see Figures \ref{fig:nao2-failure-rates}, \ref{fig:nao4-failure-rates}, \ref{fig:kuka2-failure-rates}, \ref{fig:kuka4-failure-rates}, \ref{fig:kuka7-failure-rates}). 
Note that it is not feasible to use the same value for all experiments since the initial performance across experiments varies between $\approx$50\% and $\approx$90\%. 

\begin{figure}[!tb]
	\centering
	\includegraphics[width=\columnwidth]{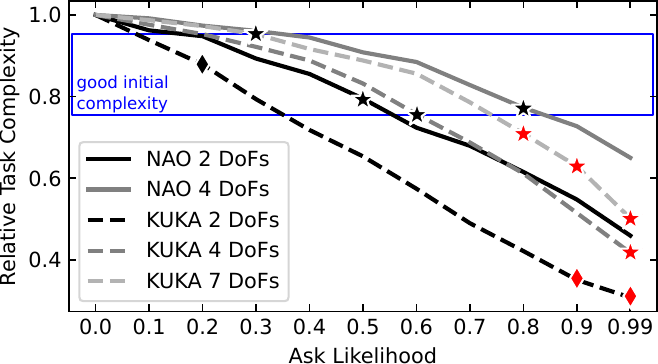}
	\caption{Relative task complexity reduction induced by feedback frequencies for all experiments. 
	Interaction has a differential effect on the complexity reduction, depending on the task. Black 	symbols show the best choice of $\mathcal{L}$ for all five experiments during early training (compare Figs. \ref{fig:nao2-failure-rates} to \ref{fig:kuka7-failure-rates}, second failure rate thresholds). Red symbols show significantly detrimental choices for $\mathcal{L}$ at this failure rate threshold.}
	\label{fig:initial-al}
\end{figure}

Figure \ref{fig:initial-al} shows the relative task complexity for all experiments and $\mathcal{L}$ values, along with the best choices of $\mathcal{L}$ at the mentioned thresholds and the significantly detrimental choices. 
Indeed, $\mathcal{L}$ values that lead to relatively low task complexities are prone to have a detrimental effect. In contrast, the most beneficial choices of early $\mathcal{L}$ values are those that lead to a relative task complexity between $\approx 0.78$ to $\approx 0.95$. 
Thus, drawing an initial $\mathcal{L}$ from that range for each task makes it considerably more likely to choose the initially optimal $\mathcal{L}$ value than naively sampling from the range of $\mathcal{L}=0.2$ to $\mathcal{L}=0.8$. 

Based on this observation, we argue against the claims that either high, intermediate, or low $\mathcal{L}$ values are optimal in early training. 
Instead, optimality seems to be better predicted by the task complexity reduction induced by $\mathcal{L}$.

\section{Conclusion}
\label{sec:conclusion}
In this study, we conducted a thorough extension of previous research investigating the effect of feedback frequency on agent performance in continuous action and state spaces. 
The main contribution is the discovery that task complexity and performance threshold influence the interpretation of the best interaction rate with a teacher. Moreover, no single best solution exists across task conditions. 

Our results instead suggest that the optimal interaction rate changes over time and that the task complexity determines the specific optimal trajectory for $\mathcal{L}$. These observations allow us to consolidate previously contradictory claims on the optimal interaction frequency.
Furthermore, we described a heuristic to choose the initial feedback frequency based on a measure of the relative task complexity changes induced by the teacher's feedback on an untrained agent.

\paragraph{Future work:}
A future goal is to probe the described heuristic further to predict the optimal trajectory before -- and potentially adjust it during -- training. Such a strategy has the potential to increase data efficiency significantly. 
Drawing such conclusions across an even more comprehensive range of tasks requires more extensive experimentation with more tasks of different complexity. 

It is also necessary to determine the deeper cause of seemingly differential effects of task complexity reduction by teacher interaction and the relation to potential capacity loss in the agents' neural network function appropriators.

Finally, it will be helpful to quantify the interaction of the feedback frequency effects with other factors, such as fixed feedback budgets and advice quality, to go towards applicable scenarios with realistic human feedback.

\backmatter
\small
\bmhead{Supplementary information} It can be found in the following link: \href{https://doi.org/10.6084/m9.figshare.20027582}{DOI: 10.6084/m9.figshare.20027582}.

\bmhead{Acknowledgments} Not applicable.

\section*{Declarations}

\bmhead{Funding} Open Access funding provided by the Projekt DEAL (Open access agreement for Germany). Research funding by the \textit{\href{https://www.dfki.de/en/web/research/projects-and-publications/projects-overview/project/m-rock}{M-RoCK -- Human-Machine Interaction Modeling for Continuous Improvement of Robot Behavior}} project funded by the Federal Ministry of Education and Research with grant no. 01IW21002.
\bmhead{Conflict of interest/Competing interests} The authors declare that they have no conflict of interest.
\bmhead{Ethics approval} This article does not contain any studies with human participants or animals performed by any of the authors.

\bmhead{Open Access} This article is licensed under a Creative Commons Attribution 4.0 International License, which permits use, sharing, adaptation, distribution and reproduction in any medium or format, as long as you give appropriate credit to the original author(s) and the source, provide a link to the Creative Commons licence, and indicate if changes were made. The images or other third party material in this article are included in the article's Creative Commons licence, unless indicated otherwise in a credit line to the material. If material is not included in the article's Creative Commons licence and your intended use is not permitted by statutory regulation or exceeds the permitted use, you will need to obtain permission directly from the copyright holder. To view a copy of this licence, visit \url{https://creativecommons.org/licenses/by/4.0/}.

\bmhead{Consent to participate} Not applicable
\bmhead{Consent for publication/Informed consent} Not applicable
\bmhead{Availability of data and materials} Not applicable


\bibliography{references.bib}

\bmhead{Publisher's Note} Springer Nature remains neutral with regard to jurisdictional claims in published maps and institutional affiliations.


\end{document}